\documentclass[conference]{IEEEtran}
\IEEEoverridecommandlockouts
\usepackage{subcaption}
\usepackage{calc}
\usepackage{amssymb}
\usepackage{algorithm}
\usepackage{algpseudocode}
\usepackage{dirtytalk}
\usepackage[hidelinks]{hyperref}
\usepackage{graphicx}
\usepackage{cite}
\usepackage{amsmath,amssymb,amsfonts}
\usepackage{graphicx}
\usepackage{textcomp}
\usepackage{xcolor}    
\usepackage{authblk}

\begin{document}

\title{Gaze-Vector Estimation in the Dark with Temporally Encoded Event-driven Neural Networks\\

\thanks{Abeer Banerjee is financially supported by CSIR-Human Resource Development Group (HRDG), Government of India, with a Senior Research Fellowship (GATE).}
}

% \author{\IEEEauthorblockN{1\textsuperscript{st} Given Name Surname}
% \IEEEauthorblockA{\textit{dept. name of organization (of Aff.)} \\
% \textit{name of organization (of Aff.)}\\
% City, Country \\
% email address or ORCID}
% \and
% \IEEEauthorblockN{2\textsuperscript{nd} Given Name Surname}
% \IEEEauthorblockA{\textit{dept. name of organization (of Aff.)} \\
% \textit{name of organization (of Aff.)}\\
% City, Country \\
% email address or ORCID}
% \and
% \IEEEauthorblockN{3\textsuperscript{rd} Given Name Surname}
% \IEEEauthorblockA{\textit{dept. name of organization (of Aff.)} \\
% \textit{name of organization (of Aff.)}\\
% City, Country \\
% email address or ORCID}
% \and
% \IEEEauthorblockN{4\textsuperscript{th} Given Name Surname}
% \IEEEauthorblockA{\textit{dept. name of organization (of Aff.)} \\
% \textit{name of organization (of Aff.)}\\
% City, Country \\
% email address or ORCID}
% \and
% \IEEEauthorblockN{5\textsuperscript{th} Given Name Surname}
% \IEEEauthorblockA{\textit{dept. name of organization (of Aff.)} \\
% \textit{name of organization (of Aff.)}\\
% City, Country \\
% email address or ORCID}
% \and
% \IEEEauthorblockN{6\textsuperscript{th} Given Name Surname}
% \IEEEauthorblockA{\textit{dept. name of organization (of Aff.)} \\
% \textit{name of organization (of Aff.)}\\
% City, Country \\
% email address or ORCID}
% }

\author[1,2]{Abeer Banerjee}
\author[1,2]{Naval K. Mehta}
\author[1,2]{Shyam S. Prasad}
\author[1,2]{Himanshu}
\author[1,2]{Sumeet Saurav}
\author[1,2]{Sanjay Singh}
\affil[1]{CSIR-Central Electronics Engineering Research Institute (CSIR-CEERI), India}
\affil[2]{Academy of Scientific and Innovative Research (AcSIR), India}

% \author{Abeer Banerjee, , , , , and }
% \date{%
%     $^1$Organization 1\\%
%     $^2$Organization 2\\[2ex]%
%     \today
% }

% \author{\IEEEauthorblockN{Abeer Banerjee}
% \and
% \IEEEauthorblockN{Naval Kishore Mehta}
% \and
% \IEEEauthorblockN{Shyam Sunder Prasad}
% \and
% \IEEEauthorblockN{Himanshu Kumar}
% \and
% \IEEEauthorblockN{Sumeet Saurav}
% \and
% \IEEEauthorblockN{Sanjay Singh}
% }

\maketitle

\begin{abstract}
In this paper, we address the intricate challenge of gaze vector prediction, a pivotal task with applications ranging from human-computer interaction to driver monitoring systems. Our innovative approach is designed for the demanding setting of extremely low-light conditions, leveraging a novel temporal event encoding scheme, and a dedicated neural network architecture. The temporal encoding method seamlessly integrates Dynamic Vision Sensor (DVS) events with grayscale guide frames, generating consecutively encoded images for input into our neural network. This unique solution not only captures diverse gaze responses from participants within the active age group but also introduces a curated dataset tailored for low-light conditions. The encoded temporal frames paired with our network showcase impressive spatial localization and reliable gaze direction in their predictions. Achieving a remarkable 100-pixel accuracy of 100\%, our research underscores the potency of our neural network to work with temporally consecutive encoded images for precise gaze vector predictions in challenging low-light videos, contributing to the advancement of gaze prediction technologies.
\end{abstract}
\begin{IEEEkeywords}
Gaze Estimation, Neuromorphic Camera, Neural Networks, Low-light, Event Dataset
\end{IEEEkeywords}
\section{Introduction}\label{sec:intro}
\begin{figure}[h]
    \centering
    \includegraphics[width=0.48\textwidth]{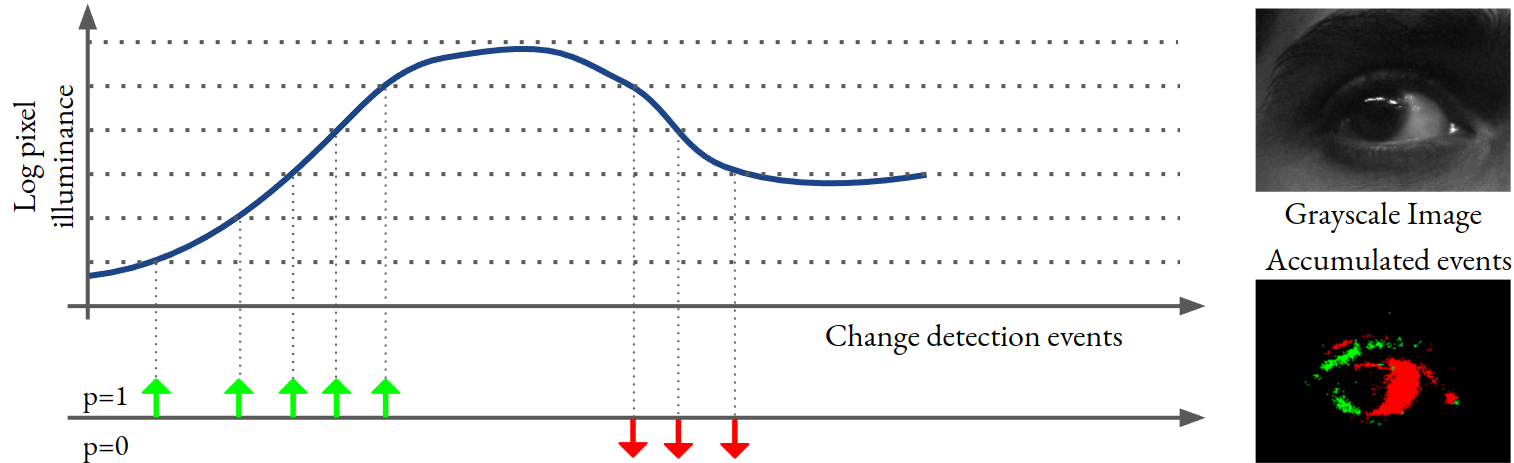}  
    \caption{Working principle of an event-based sensor \cite{banerjee2022gaze}. The change in events is denoted in terms of polarities.}
    \label{dvs}
\end{figure}

The ability to predict the human gaze plays a pivotal role in understanding cognitive processes, human-computer interaction, and various applications in fields such as psychology, neuroscience, and technology. Gaze prediction, particularly the anticipation of saccadic eye motion, has garnered significant attention due to its potential to decipher human intent, preferences, and decision-making patterns. In low light, the human eye undergoes physiological changes that significantly impact the predictability of saccades. The interplay between ambient light levels, pupil dilation, and ocular movements adds layers of complexity to the already intricate task of gaze prediction. Addressing the research gaps in gaze prediction, particularly in low light conditions, necessitates not only innovative hardware solutions but also sophisticated data recording techniques and advanced algorithms. The efficacy of any prediction model relies heavily on the quality and diversity of training data. However, the scarcity of comprehensive datasets capturing saccadic eye motion in low light conditions impedes progress. Researchers grapple with the challenge of acquiring and curating datasets that mirror the complexities of real-world scenarios, limiting the robustness of gaze prediction systems. The utilization of event-based camera systems is increasingly favored for such tasks, owing to their low latency and extensive dynamic range, as discussed in \cite{cheng2019det}.  
\begin{figure}[t]
\centering
\includegraphics[width=0.495\textwidth]{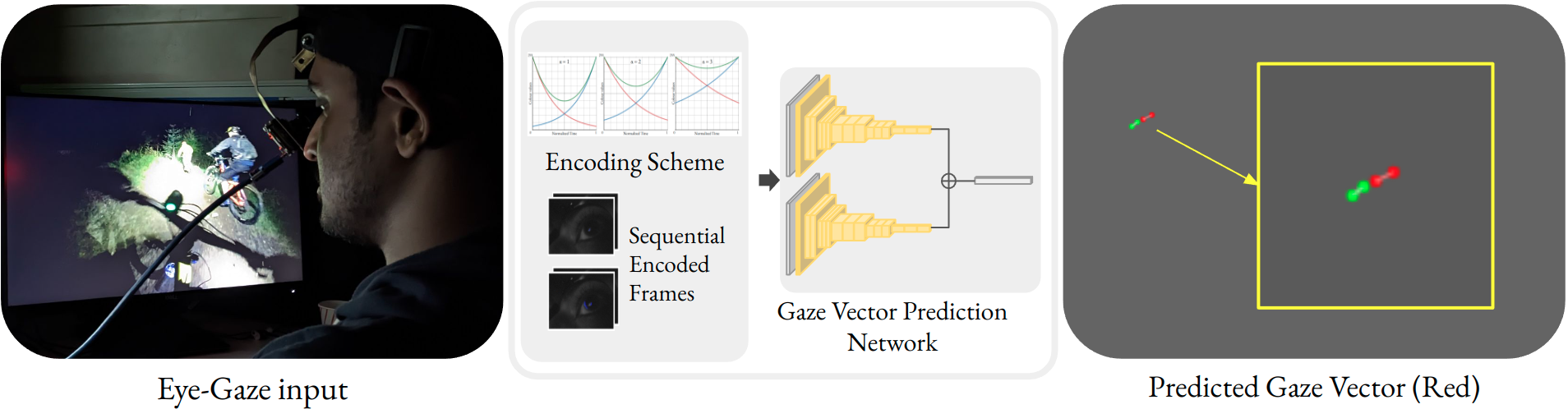}  
\caption{Gaze Vector prediction (red) in an extremely light-starved condition obtained using our method.}
\label{tag}
\end{figure}
Previous studies have delved into the utilization of events for near-eye gaze detection under controlled scenarios. The working principle of an event-based sensor is highlighted in Fig. \ref{dvs}. 
\begin{figure*}[t]
    \centering
    \includegraphics[width=1\textwidth]{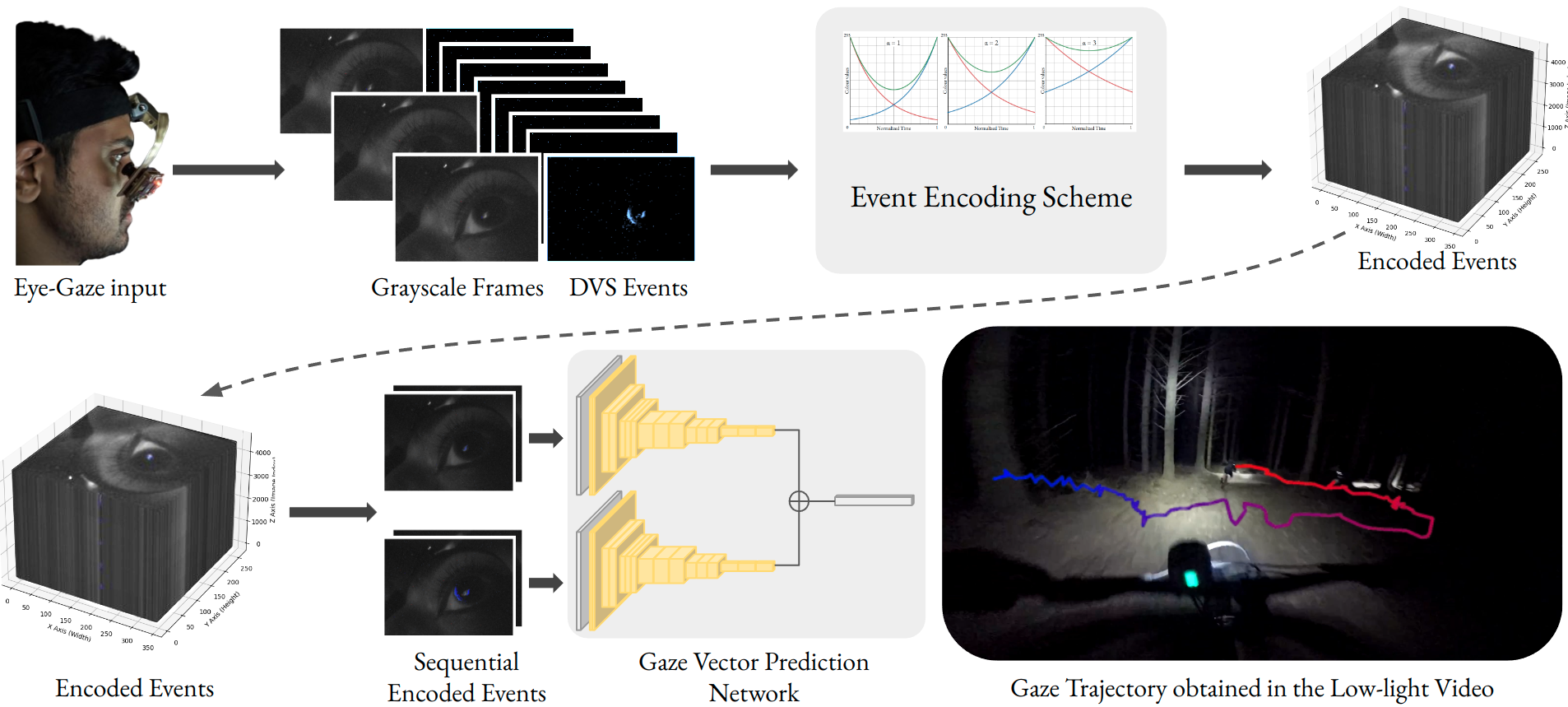}  
    \caption{Overview of our gaze vector prediction pipeline in extremely light-starved situations: The eye-gaze input from a head-mounted DVS camera is temporally encoded to form a stacked representation of fused grayscale frames and events. Consecutive temporally encoded frames are set as inputs to the Gaze Vector Prediction Neural Network. The final gaze trajectory is plotted on the low-light video frame where the transition from blue to red color depends on the arrival time of the events.}
    \label{overv}
\end{figure*}
Gaze tracking involves estimating and tracking a person's eye movement to achieve a specific goal. Eye movement has long served as a signal for interacting with computers \cite{Kols_2017}. Advancements include developing assistance systems for disabled individuals, eliminating hand dependency for computer use, and introducing the eye-mouse, which responds to eye movement \cite{https://doi.org/10.48550/arxiv.1609.07342}. Furthermore, gaze tracking finds application in interactive domains like virtual learning, computer gaming \cite{6227433}, and augmented reality/virtual reality. Noteworthy use cases extend to medicine and healthcare, such as diagnosing concussion and progressive neurological disorders \cite{Bruny2019ARO}, particularly in cases where patients face mobility challenges.

Earlier methodologies for eye-tracking encompassed electro-oculography, search coils, and various unique approaches \cite{Young1975SurveyOE}. CNN-based approaches \cite{Bao_2022,7299081,8122058}, trained on extensive datasets like MPIIGaze \cite{8122058}, have been pivotal in estimating eye gaze direction. Camera-based eye-tracking systems have evolved from Purkinje reflection-based techniques \cite{Cornsweet:73, Crane:85} to model-based approaches that track eye movement by extracting frames from video sequences \cite{LI20101377, 840620, Wang2017RealTE}. Angelopoulos et al. \cite{9389490} introduced a hybrid model by combining events from an event-based camera with frames obtained from a standard camera. The introduction of a bio-inspired camera system, i.e., a neuromorphic event camera has revolutionized the vision research community and these have been employed for a variety of challenging tasks including fall detection, gesture recognition \cite{prasad2022real, prasad2023hybrid, amir2017low}.  Frame-based details initialize pupil fitting functions, and event information supplements frame details, enabling gaze estimation. However, the model's reliance on frames renders it susceptible to motion blur. In response, \cite{9706617} proposes a method compatible with event cameras and flash lighting conditions. They amplify specular components via coded differential lighting, creating corneal glints.
Our contributions are in three fundamental aspects:
\begin{itemize}
\item Gaze vector prediction with a designed neural network architecture for handling consecutive encoded frames obtained from grayscale images and DVS events. Parallel integration of pre-trained encoders for feature extraction and prediction of temporally consecutive centroids for accurate person-independent gaze estimation.
\item A novel temporal encoding scheme for combined processing of event data and grayscale guide frames by sequential stacking of DVS events, and fusion with nearest grayscale frame in a time synchronous way.
\item A new event-based eye-gaze dataset captured in extreme low-light conditions, requiring active subjects to rapidly track nighttime biking POV videos where traditional frame-based methods struggle.
\end{itemize}
In the subsequent sections of this paper, we outline the dataset collection process, encoding pipeline, network architecture, and training methodology. We also present both qualitative and quantitative evaluation results, affirming the effectiveness of our approach.

\section{Overview}
Fig. \ref{overv} illustrates the overview of our approach. The dataset, recording spike events paired with grayscale frames, and our robust neural network collectively navigate complexities posed by low light, saccadic eye motion, and rapid gaze changes. The subsequent sections dive into more detail regarding the intricacies of our research.
\section{Methodolgy}
\subsection{Data Collection} 
This research introduces a dataset, Gaze-FELL: Gaze Frames and Events in Low-Light. This dataset is for gaze detection, particularly in extreme low-light conditions, and was captured from five healthy subjects aged 25 to 35. The experiments utilized a head-mounted Dynamic Vision Sensor (DAVIS 346) to record the eye gaze of the subjects and the results grayscale frames and events were stored. The intentional low-light setup explores gaze behavior where traditional methods falter. Subjects tracked targets in 14 nighttime biking videos, simulating scenarios demanding swift and accurate gaze tracking under challenging conditions. The dataset will be made public on request. 
\begin{figure}[t]
    \centering
    \includegraphics[width=0.48\textwidth]{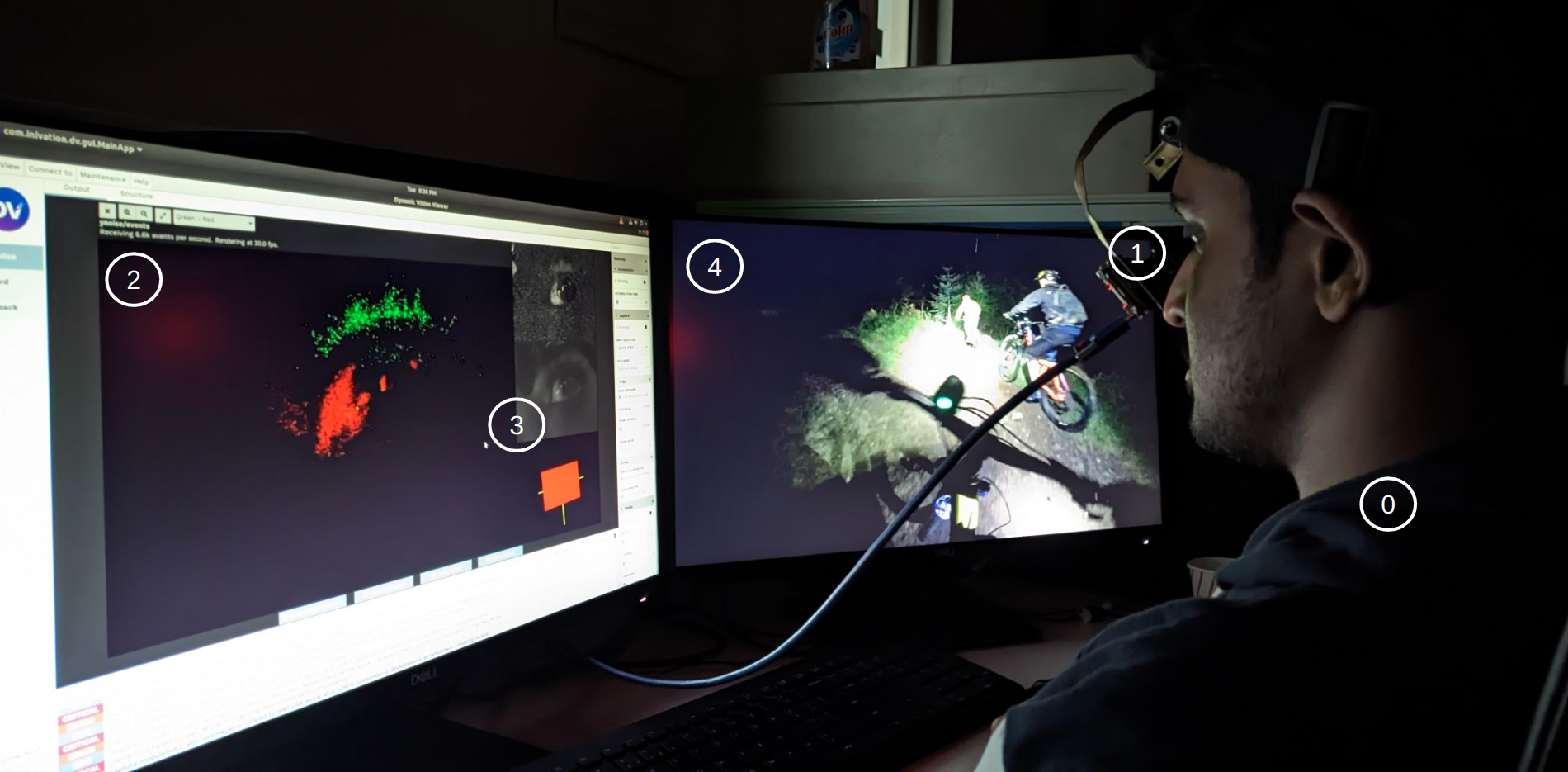}  
    \caption{This is a representational version of the low-light recording setup. (0) Subject (1) DVS mounted on headgear (2) DVS events (3) Grayscale frames (4) Video stimulus.}
    \label{rec}
\end{figure}
In low-light conditions, saccadic eye movements are essential, reflecting rapid and involuntary shifts in gaze toward points of interest. Analyzing saccades in challenging environments is crucial for understanding gaze adaptation to low-light scenarios. The proposed dataset uniquely tackles frame-based method limitations in extreme low-light conditions, providing a valuable resource for robust gaze detection systems. Fig. \ref{rec} illustrates our low-light recording setup, where participants viewed a single low-light video stimulus with the screen as the sole illumination source during data collection.

\begin{figure}[h]
    \centering
    \includegraphics[width=0.48\textwidth]{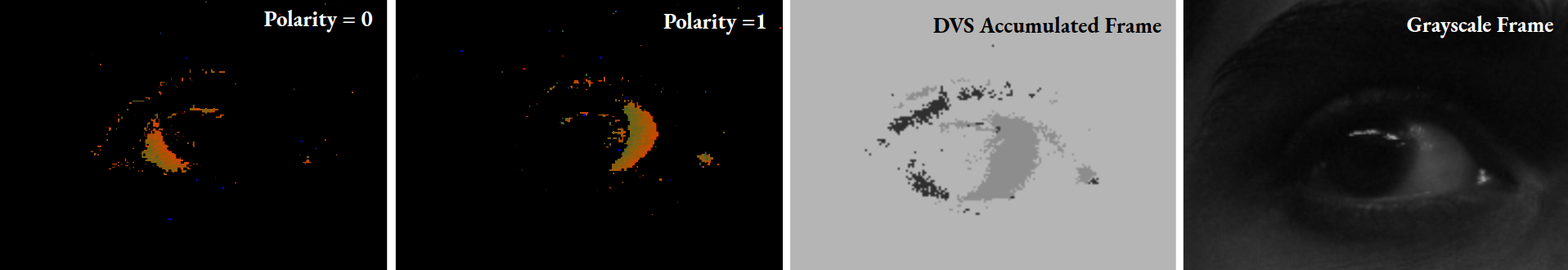}
    \caption{Event-Encoded images obtained using the encoding functions in contrast to DVS accumulated image.}
    \label{fig_encode}
\end{figure}

\textbf{Event Encoding:} Our method uses a rate-coded event-to-image encoding technique for eye-gaze estimation \cite{banerjee2022gaze}. The concept of employing an event-to-image encoding scheme stems from the utilization of Convolutional Neural Networks for regression tasks. In instances of rapid eye movements, events occur sporadically, leading to the creation of sparse event logs. Introducing a compression technique that encodes the temporal rate of information change proves beneficial for visualizing comprehensive gaze events. As a result, timestamps are normalized to capture and represent the rate of change.
\begin{equation}
r = e^{- \alpha t} \quad
b = e^{\alpha (t-1)} \quad
g =  e^{- \alpha t} + e^{\alpha (t-1)} - \frac{1}{e^\alpha}
\label{encode}
\end{equation}
In the image domain, we use color values to distinguish between different event samples from different points in time normalized within the range of $t \in [0,1]$. We use functions for each color channel according to \cite{banerjee2022gaze} as described in Eq. \ref{encode} and refer to this encoding function as $\eta_f$ in this paper.  The images computed using the encoding function have been plotted in Fig. \ref{fig_encode} and the encoding algorithm is elaborated in Alg. \ref{alg:enc}. 
\begin{algorithm}[h]
 \caption{Event encoding function: $\eta_f$}\label{alg:enc}
 \begin{algorithmic}
 \State $  \textbf{E} \gets \{(x_{k},y_{k},t_{k},p_{k}) \} $  \Comment{ \textbf{Events}}
  \State $  \textbf{T} \gets \{t_{1},t_{2},....,t_{n} \} $   \Comment{\textbf{Absolute Timestamps}} \\
 \State $  \textbf{\~{T}} \gets \frac{ t - t_{MIN}}{ t_{MAX} - t_{MIN}}   $ \Comment{\textbf{Normalized Timestamps}} \\
  \State $  \textbf{{I}} \gets \mathbb{R}^{H \times W \times C }  $ \Comment{\textbf{Blank Frame}}

\For {$t \hspace{2mm}  \textbf{in} \hspace{2mm} \textbf{\~{T}}$}
 \If { $ \hspace{2mm} p_{t}   \hspace{2mm} is  \hspace{2mm} 0$}
 \State $ \hspace{2mm} \textbf{I(x,y,0:3)} \gets   \hspace{2mm} \textbf{ $r[t], g[t], b[t]$ } $ 
% % \EndIf
 \Else
 \State $ \hspace{2mm} \textbf{I(x,y,3:6)} \gets   \textbf{ $r[t], g[t], b[t]$ } $ 
\EndIf
\EndFor
\State \bf{return} \textbf{I} 
\end{algorithmic}
\end{algorithm}
% \begin{algorithm}[h]
%  \caption{Event encoding function: $\eta_f$}\label{alg:enc}
%  \begin{algorithmic}
%  \State $  \textbf{E} \gets \{(x_{k},y_{k},t_{k},p_{k}) \} $  \Comment{ \textbf{Events}}
%   \State $  \textbf{T} \gets \{t_{1},t_{2},....,t_{n} \} $   \Comment{\textbf{Absolute Timestamps}} \\
%  \State $  \textbf{\~{T}} \gets \frac{ t - t_{MIN}}{ t_{MAX} - t_{MIN}}   $ \Comment{\textbf{Normalized Timestamps}} \\
%   \State $  \textbf{{I}} \gets \mathbb{R}^{H \times W \times C }  $ \Comment{\textbf{Blank Frame}}

% \For {$t \hspace{2mm}  \textbf{in} \hspace{2mm} \textbf{\~{T}}$}
%  \If { $ \hspace{2mm} p_{t}   \hspace{2mm} is  \hspace{2mm} 0$}
%  \State $ \hspace{2mm} \textbf{I(x,y,0:3)} \gets   \hspace{2mm} \textbf{ $r[t], g[t], b[t]$ } $ 
% % % \EndIf
%  \Else
%  \State $ \hspace{2mm} \textbf{I(x,y,3:6)} \gets   \textbf{ $r[t], g[t], b[t]$ } $ 
% \EndIf
% \EndFor
% \State \bf{return} \textbf{I} 
% \end{algorithmic}
% \end{algorithm}

\begin{figure*}[t]
    \centering
    \includegraphics[width=1\textwidth]{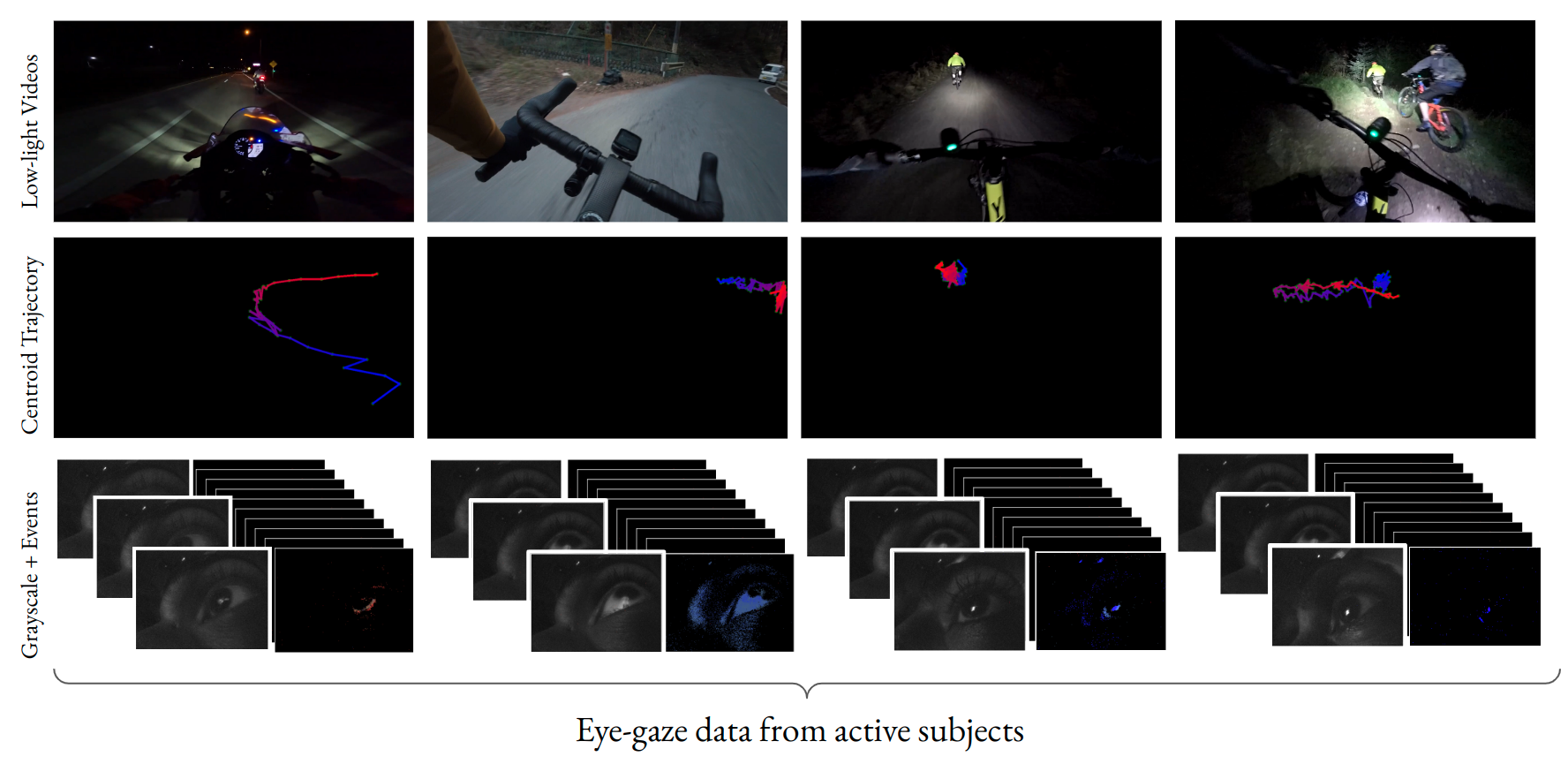}  
    \caption{Gaze-FELL dataset: The low-light videos are in first-person view, and the gaze trajectory is plotted below each video. The encoded frame and event data for some of the subjects have been illustrated below each gaze trajectory.}
    \label{dataset}
\end{figure*}
\subsection{Temporal Encoding}
This research presents an innovative event-to-image encoding technique designed for accurate gaze detection in extreme low-light conditions. The encoding process involves capturing Dynamic Vision Sensor (DVS) events in temporal bins of 33 milliseconds, coupled with the fusion of these events with the nearest corresponding grayscale frames in time. Ground truth centroids, essential for model training, are obtained through manual annotation by subjects who track points of interest during short video clips. The events are temporally binned in intervals of 33 milliseconds, denoted as $T_{bin}$, forming discrete temporal units. Grayscale frames are captured at a lower frame rate, approximately 2 to 3 frames per second (FPS), denoted as $F_{gray}$. The temporal resolution mismatch between DVS events and grayscale frames necessitates a careful synchronization strategy. Note that the DVS events and the grayscale frames are captured using the same device, i.e., the DAVIS 346 DVS event camera. In each iteration, $n$, the encoding process involves aggregating events within the temporal bin $T_{bin}$ to create a compact representation $E_{n}$. The fusion with corresponding grayscale frames is performed by pairing $E_n$ with the nearest $F_{gray}$ frame in time, denoted as $F_n$, as shown in Fig. \ref{dataset}. The equations representing the temporal encoding process are shown in Eq. \ref{eqen}, and Eq. \ref{eqfn}. 
\begin{equation}
\label{eqen}
E_n = \sum_{t_{k} = t_0}^{n \cdot T_{bin} + t_0} [x_k, y_k, t_k, p_k]
\end{equation}
\begin{equation}
\label{eqfn}
F_n = arg\: min_{F_{gray}} \lvert T_{E_{n}} - T_{F_{gray}} \rvert
\end{equation}

The temporal encoding results in more than 8,500 paired images and their corresponding gaze centroids.
\begin{figure*}[t]
    \centering
    \includegraphics[width=1\textwidth]{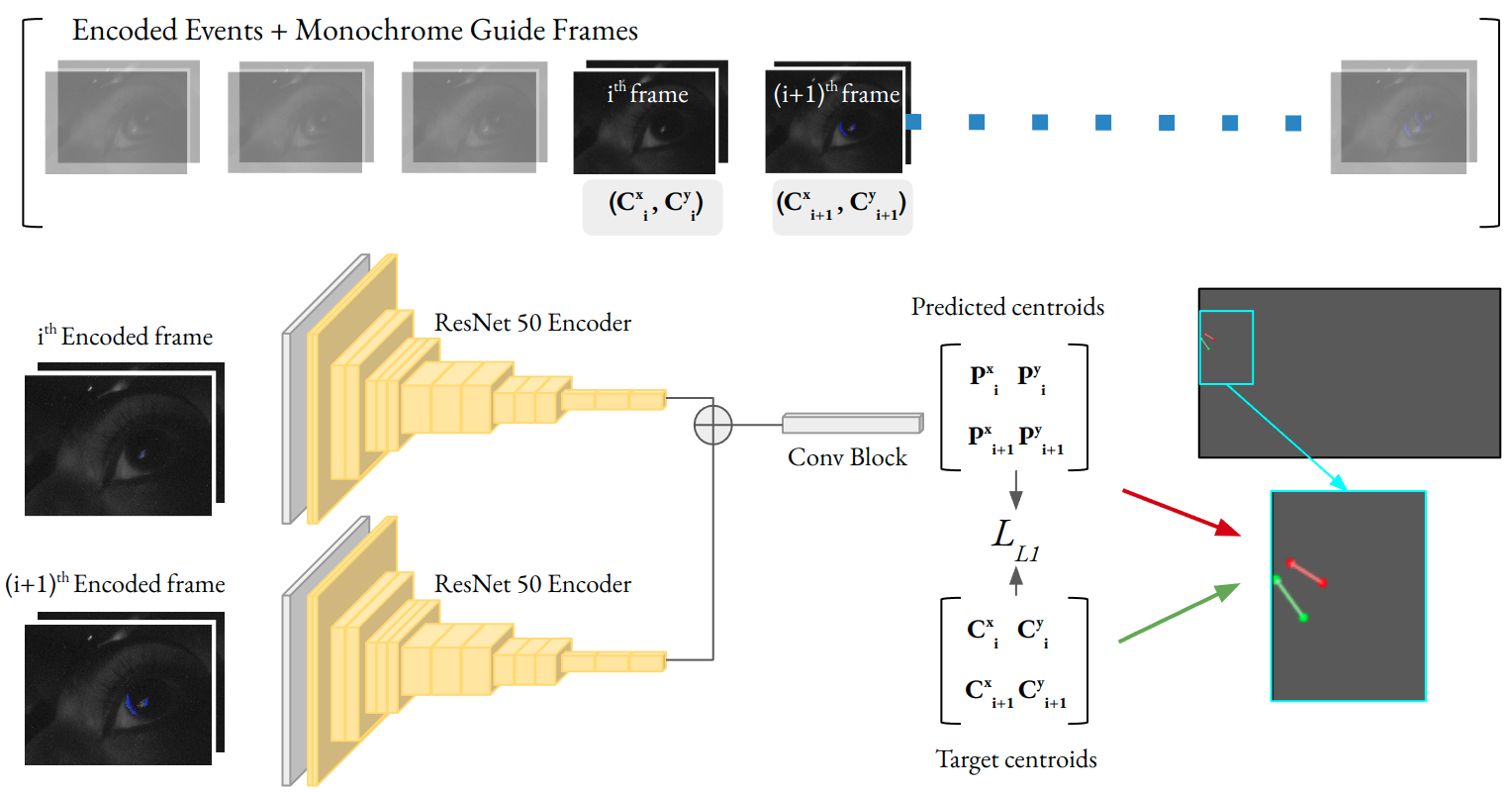}  
    \caption{This is the training pipeline of our gaze vector prediction approach. The consecutive temporally encoded frames are set as inputs to our designed network architecture consisting of two truncated ResNet-50 feature extractors. The predicted pair of centroids form the red gaze vector and the target centroids form the green vector. The L1-loss ($L_{L1}$) calculated from this predicted and target centroids are backpropagated through the network.}
    \label{train}
\end{figure*}
The algorithm, Alg. \ref{alg:framing}, is for temporal encoding of the events and grayscale frames. The frames thus obtained are paired up with the manually annotated centroids obtained from the reference video. It operates on a set of frames denoted by the variables $N_V$, and $N_G$. For each frame in the set $N_V$, the algorithm repeats the grayscale frame and initializes a start time variable ($start$). For each iteration, it determines the end time ($end$) by adding multiples of the temporal difference between video frames ($T_V$) to the start time. 
Using these time indices, it extracts a subset of timestamps ($t$) from the original timestamp array ($T$). If the length of $t$ is greater than 1, indicating the presence of events, it utilizes the color encoding function ($\eta_f$), to produce rate-coded events with varying colors. The encoded image provides a visual representation of the event encoding process. The algorithm returns the encoded frames paired up with the corresponding ground truth centroid from the list $C$. The $N_{loop}$ variable keeps track of the current iteration number and is used to pair up the final frames with the ground truth centroids. The ground truth centroids are obtained from the $C$, which contains manually annotated centroid points corresponding to each frame in the reference video. This algorithm uses temporal information from DVS events with spatial information from grayscale frames, forms encoded images, and pairs them with the corresponding ground truth centroids.
\subsection{Network Architecture} Our network performs a temporal fusion of information from encoded grayscale and event frames as shown in Fig. \ref{train}. It leverages two consecutive temporally encoded frames as input to predict consecutive gaze centroids. 
\begin{algorithm}[h]
\caption{Event and Frame Fusion Algorithm}\label{alg:framing}
\begin{algorithmic}
\State $  \textbf{E} \gets \{(x_{k},y_{k},t_{k},p_{k}) \} $  \Comment{ \textbf{Events}}
\State $ \textbf{T} \gets \{t_{1},t_{2},....,t_{n} \} $  \Comment{\textbf{Absolute Timestamps}}
\State $T_V \gets (T.max - T.min)/N_V$ \Comment{\textbf{$N_V$=\# GT Frames}}
\State $T_G \gets (T.max - T.min)/N_G$ \Comment{\textbf{$N_G$=\# Gray Frames}}
\State $N_{loop} \gets 0$  
\For{$j$ \textbf{in} $N_G$} 
\State $I_{enc} \gets F_j$ \Comment{$j^{th}$ \textbf{Gray Frame}}
\If{$j == 0$} \State $start \gets T[j]$
\Else \State $start \gets end$
\EndIf
\For{$k$ \textbf{in} $T_G/T_V$}
\State $end \gets start + k \cdot T_V $
\State $s\_index \gets (\text{argmin}(\lvert T - start \rvert))$
\State $e\_index \gets (\text{argmin}(\lvert T - end \rvert))$
\State $t \gets T[s\_index : e\_index]$
\If{$\lvert t \rvert > 1$}
\State $I_{enc} \gets \text{Fuse}(t, I_{enc}, \eta_f , x, y)$
\EndIf
\State $N_{loop} \gets N_{loop} + 1$
\State \textbf{return} $I_{enc}, C[N_{loop}]$
\EndFor
\EndFor
\end{algorithmic}
\end{algorithm}
The architecture includes two branches, each with a truncated ResNet-50 feature extractor and a convolutional block tailored for six-channel temporally encoded images. The input to our network consists of two consecutive temporally encoded frames, denoted as the $i^{th}$ and $(i+1)^{th}$ frames. These frames are obtained through the encoding process that fuses grayscale and event frames in a temporally synchronized manner, allowing the network to capture both spatial and temporal information. The ResNet-50 model is tailored to retain essential spatial features while efficiently handling the six-channel input representing temporally encoded frames. The extracted features from the two branches are concatenated to form a joint feature representation. This concatenated feature vector is further processed through a convolutional block, to predict a pair of coordinates that form the gaze vector.

\subsection{Loss functions and Training}
The final layer of the network predicts consecutive gaze centroids based on the encoded fused feature representation. The predicted gaze vector is then compared with the target gaze vector formed using the manually annotated consecutive centroids. $L_{centroid}$ is calculated for each centroid pair, $C_i = (C_i^x , C_i^y)$ and $P_i = (P_i^x , P_i^y)$, serving as the objective function during training. We also minimize the angle formed by the predicted gaze vector and the target gaze vector modeled using $L_{\theta}$. The equations for loss calculation have been provided in Eq. \ref{lcen}, and Eq. \ref{lang}.
\begin{equation}\label{lcen}
L_{centroid}= \frac{1}{N}\sum_{i=1}^{N} \left| C_{i} - P_{i} \right| + \frac{1}{N}\sum_{i=1}^{N} \left| C_{i+1} - P_{i+1} \right|
\end{equation}
\begin{equation}\label{lang}
L_{\theta} = \frac{1}{N} \sum_{i=1}^{N} \left| \cos^{-1}[\overrightarrow{(C_{i+1}-C_{i})} \cdot \overrightarrow{(P_{i+1}-P_{i})}] \right|
\end{equation}
The losses are backpropagated through the network to update the model parameters, optimizing its ability to accurately predict gaze vectors. We have used the Adam optimizer with a learning rate of $1e-3$ for this task. Our network architecture offers a novel approach to gaze vector prediction by integrating spatial and temporal information. The temporal fusion of temporally encoded frames ensures that the network captures dynamic gaze patterns, especially during rapid eye movements. The use of ResNet-50 feature extractors for temporally encoded data enhances the network's ability to discern crucial features contributing to accurate gaze predictions. This architecture holds promise in enhancing gaze detection accuracy, particularly in scenarios where temporal dynamics play a critical role, such as low-light conditions or dynamic visual environments. The utilization of L1 loss facilitates efficient training and convergence, ensuring that the network learns to predict consecutive gaze centroids effectively. 

\section{Results}
\begin{figure}[h]
    \centering
    \includegraphics[width=0.48\textwidth]{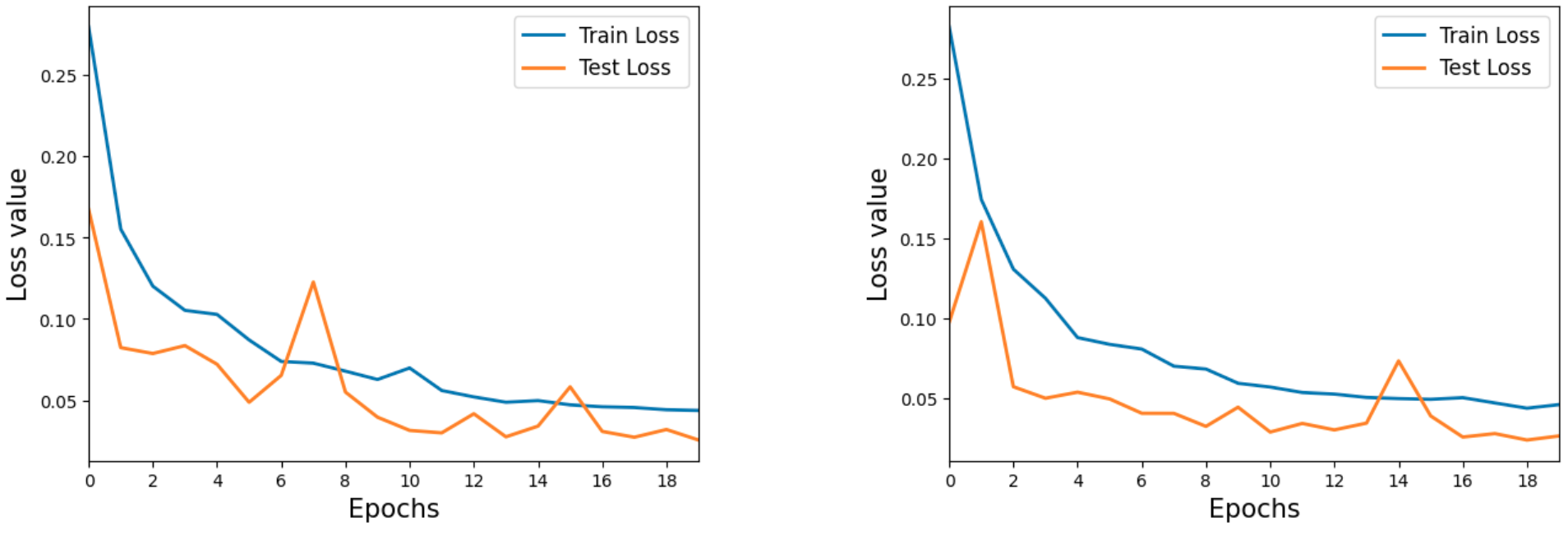}  
    \caption{Training and testing loss curves: The first plot is for $L_{centroid}$ only while the second plot is for the total loss of $L_{centroid} + L_{\theta}$.}
    \label{loss}
\end{figure}

Our evaluation includes both quantitative and qualitative assessments of the proposed gaze detection model. The inclusion of training and testing curves provides insights into the model's learning dynamics, while qualitative results offer a visual interpretation of its predictive capabilities. The training illustrates the convergence of the model, indicating that the network has effectively learned to predict gaze vectors under light-starved scenarios. The testing curve showcases that the model has generalized well over multiple subjects and can adapt to unseen scenarios that involve challenging illumination constraints, and the alignment of the two curves signifies the model is robust. Fig. \ref{loss} shows that incorporating $L_{\theta}$ enables a faster and smoother convergence of the model by enforcing the predicted vectors to align well with the target vectors.

\begin{figure}[h]
    \centering
    \includegraphics[width=0.48\textwidth]{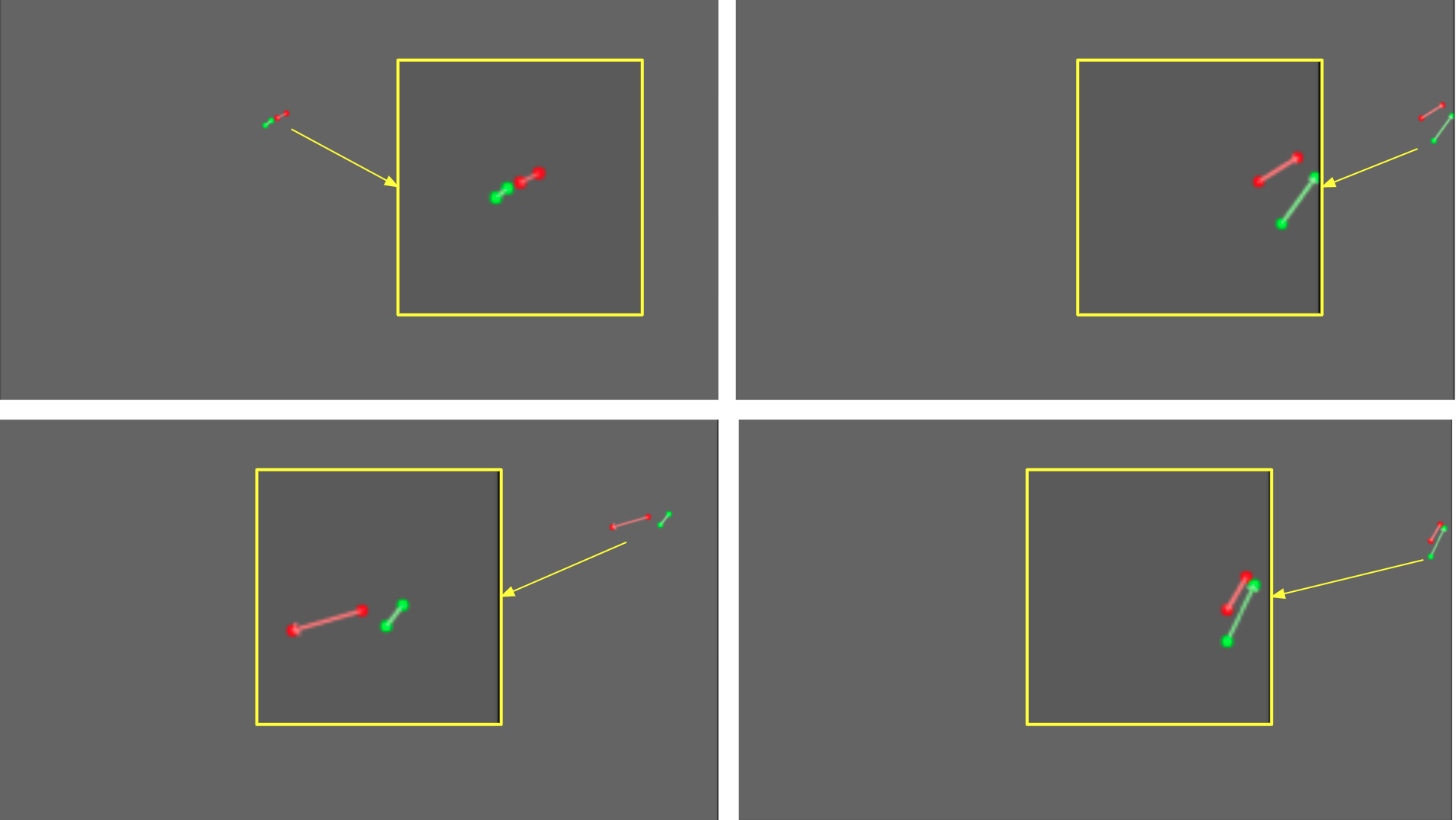}  
    \caption{The above figure shows the gaze vector prediction (in red) corresponding to tricky boundary cases of ground truths (in green). }
    \label{pred}
\end{figure}

Quantitative accuracy was assessed using two methods. First, considering a variable-radius circle centered on the target gaze vector centroid, a trial was deemed successful if centroids of the predicted vectors fell within the circle. Second, a trial succeeded if any part of the predicted vector crossed through the circle formed using the target centroid, making strategy 1 more challenging. Table \ref{acc_table} displays the localization accuracy of the prediction pipeline, demonstrating the model's proficiency in accurately capturing gaze direction. Qualitative results in Fig. \ref{pred} visually represent predicted gaze vectors, showcasing spatial localization proficiency. Minor discrepancies in direction are observed during gaze shifts, yet overall alignment with target vectors is impressive for continuous gaze tracking. 
\begin{table}[h]
\centering
\captionof{table}{Testing accuracy obtained corresponding to different radii in pixels.}
\begin{tabular}{lll} 
\hline
\textbf{Pixel Radius} & \textbf{Strat 1 Acc.(\%)} & \textbf{Strat 2 Acc.(\%)}\\
\hline
100   &  100.00 &  100.00  \\
90    &   99.27 &   100.00\\ 
75    &   94.71 &   97.35 \\
50    &   69.23 &   77.64 \\
25    &   63.94 &   69.47 \\
\hline
\end{tabular}
\label{acc_table}
\end{table}

The observed spatial localization of gaze vectors highlights the model's ability to accurately predict the subject's focal points. The minor discrepancies in gaze direction changes suggest areas for potential refinement, especially in handling abrupt changes in gaze dynamics. Nevertheless, the overall performance indicates that the proposed model holds promise for robust and continuous gaze-tracking applications.

\section{Discussion and Conclusion}
The proposed gaze vector prediction pipeline has demonstrated remarkable accuracy in predicting gaze vectors, particularly in achieving precise spatial localization. However, some challenges arise in the accurate determination of gaze direction in specific scenarios. The subtle discrepancies observed in the direction of the gaze vector occur infrequently and are primarily attributed to changes in the subject's viewpoint within the short time frame of 33 milliseconds. Addressing these rare occurrences would improve the model's robustness in subsequent iterations of the research.

In conclusion, introducing a unique dataset captured in extremely low-light conditions presents a novel avenue for eye-gaze research, emphasizing the significance of saccadic eye motion. The proposed approach tackles this challenging problem through the innovative integration of a specialized neural network and a temporal encoding scheme that leverages both Dynamic Vision Sensor (DVS) and grayscale frames. The network successfully localizes gaze vectors across the dataset, showcasing its robustness even in challenging low-light environments. The accomplishments of the proposed methodology underscore its potential contributions to advancing gaze detection in real-world scenarios, particularly those characterized by low-light conditions. The findings lay the foundation for future research in improving gaze prediction models, aiming to address the complexities associated with rapid changes in gaze direction during dynamic scenarios.
% -------------------------------------------------------------------------
\bibliographystyle{IEEEtran}
\bibliography{bib}
\end{document}